\documentclass{article}

\usepackage{microtype}
\usepackage{graphicx}
\usepackage{subcaption}
\usepackage{booktabs}
\usepackage{array}
\usepackage{multirow}
\usepackage{hyperref}
\usepackage[preprint]{icml2026}
\usepackage{amsmath}
\usepackage{amssymb}
\usepackage{mathtools}
\usepackage{amsthm}
\usepackage{algorithm}
\usepackage{algorithmic}
\usepackage[capitalize,noabbrev]{cleveref}

\newcommand{\method}{\textsc{Cartograph}}
\newcommand{\uopt}{\textsc{Cartograph-A}}
\newcommand{\scorecart}{\operatorname{score}_{\mathrm{cart}}}
\newcommand{\scoreaopt}{\operatorname{score}_{\mathrm{A}}}
\newcommand{\tr}{\operatorname{tr}}

\newcommand{\rank}{\operatorname{rank}}
\newcommand{\EIG}{\operatorname{EIG}}
\newcommand{\real}{\mathbb{R}}
\newcommand{\expect}{\mathbb{E}}
\newcommand{\norm}[1]{\left\lVert #1 \right\rVert}
\newcommand{\safeincludegraphics}[2][]{%
  \IfFileExists{#2}{\includegraphics[#1]{#2}}{%
    \fbox{\parbox[c][1.6in][c]{0.95\linewidth}{\centering Figure placeholder\\Upload \texttt{#2} to render this figure.}}%
  }%
}

\theoremstyle{plain}
\newtheorem{theorem}{Theorem}[section]
\newtheorem{proposition}[theorem]{Proposition}
\newtheorem{corollary}[theorem]{Corollary}

\theoremstyle{definition}

\theoremstyle{remark}
\newtheorem{remark}[theorem]{Remark}

\icmltitlerunning{When Should an AI Scientist Stop?}

\begin{document}

\twocolumn[
\icmltitle{When Should an AI Scientist Stop?\\
Verifiable Experiment Steering and Refusal for Autonomous Discovery}

\begin{icmlauthorlist}
\icmlauthor{Neel Tushar Shah}{iitb}
\icmlauthor{Manglam Kartik}{iitb}
\end{icmlauthorlist}
\icmlaffiliation{iitb}{Indian Institute of Technology Bombay, Mumbai, India}
\icmlcorrespondingauthor{Neel Tushar Shah}{23b4244@iitb.ac.in}
\icmlcorrespondingauthor{Manglam Kartik}{23b4243@iitb.ac.in}
\icmlkeywords{AI for Science, autonomous discovery, experimental design, model discrimination, refusal, governance}
\vskip 0.3in
]

\printAffiliationsAndNotice{}

\begin{abstract}
We present \method, a verification layer for AI scientists that couples
unresolved-subspace experiment steering (\emph{select}), explicit ambiguity
closure (\emph{resolve}), and residual-based library inadequacy detection
(\emph{refuse}). Under a local linear-Gaussian bridge, raw unresolved
projection is the isotropic unresolved Fisher-information trace, while
\uopt\ is the exact unresolved A-optimal rule; closed-form EIG and
Box--Hill arise as local comparators rather than global equivalents. Across
five testbeds, \uopt\ beats raw projection
$129\mathrm{W}/0\mathrm{T}/15\mathrm{L}$ at $d\!=\!8$ ($p<10^{-21}$) in a
replicated structured cascade. More distinctively, the framework tentatively
identifies three out-of-library pharmacokinetic mechanisms and then
\emph{revokes} those identifications as residuals expose structural misfit,
while one perturbed in-library control stays identified throughout. In
low-dimensional pharmacokinetic and filtered EPA settings, near-ties against
disagreement are predicted by theory and honestly observed. Finally, in a
retrospective audit of $40$ positive claims from the published A-Lab
autonomous materials system, the refuse guard flags all $4$ claims later
marked inconclusive under manual re-analysis while passing $32/36$ confirmed
claims.
\end{abstract}

\section{Introduction}

AI systems are now participating in closed scientific discovery loops: LLM
planners propose experiments, automated laboratories execute them, and
statistical or neural modules interpret the data
\citep{king2009robot,burger2020mobile,boiko2023coscientist,szymanski2023alab,lu2024aiscientist,bran2023chemcrow,wang2023scientific}.
Complementary work has produced end-to-end automated scientific
capabilities---high-accuracy protein structure prediction
\citep{jumper2021alphafold}, scaled materials discovery
\citep{merchant2023gnome}, and symbolic-equation discovery from data
\citep{schmidt2009distilling,brunton2016sindy,udrescu2020aifeynman,cranmer2020pysr}---but
none of these stacks emits a verifiable \emph{refuse} signal when
the library or hypothesis space is structurally inadequate.
In this regime, the bottleneck is no longer \emph{proposing} experiments---LLMs
will produce more candidates than the lab can run---but deciding
\emph{which} experiment is genuinely informative, \emph{when} the current
mechanistic question is resolved, and \emph{when} the system should stop
making claims altogether because the model library it is searching is
structurally wrong.

This paper frames the verification-and-steering layer of an AI scientist as
three linked decisions:

\begin{description}
\item[Select.] Which candidate experiment most directly reduces the currently
unresolved scientific ambiguity?
\item[Resolve.] When is the ambiguity small enough to declare a mechanistic
question answered?
\item[Refuse.] When should the system stop identifying any model in the
current library because the library itself is inadequate?
\end{description}

Existing modern Bayesian experimental design (BOED)
\citep{chaloner1995review,ryan2016review,rainforth2024modern,foster2019variational,kleinegesse2020mine,blau2022rl,foster2021deep}
answers \emph{select} thoroughly. Classical model discrimination criteria
\citep{boxhill1967,atkinsonfedorov1975,pukelsheim2006optimal} answer a
restricted form of \emph{select} and \emph{resolve}. Neither literature makes
\emph{refuse} a first-class output: BOED assumes the prior support contains
the truth; model-discrimination assumes at least one rival is correct. For
high-stakes autonomous discovery---clinical pharmacokinetics, materials
synthesis, toxicology---this gap matters.

\paragraph{Contributions.}
We contribute (i)~a formal \emph{access-model} distinction separating
symbolic and behavioral querying of a scientific library, with a clean
coverage-vs-rank recovery characterization; (ii)~\method, a
verification-and-steering layer that combines unresolved-subspace steering
for \emph{select}/\emph{resolve} with a residual- and gap-based
\emph{refuse} guard;
(iii)~a local BOED bridge: exact unresolved Fisher-information identity,
exact $k=1$ posterior-variance equivalence, and first-order links to
closed-form EIG together with an isotropic-limit reduction of Box--Hill;
(iv)~an exact random-candidate scaling law that
\emph{predicts} when unresolved-subspace methods will merely tie simple
disagreement heuristics and when they will dominate; (v)~empirical evidence
across a symbolic dynamical system, a scalable structured nonlinear cascade,
a pharmacokinetic model-library benchmark, public EPA real-data time
series, and a retrospective audit of A-Lab autonomous-materials claims,
including a strongly replicated $d\!\in\!\{2,4,8,16\}$ cascade
($p<10^{-21}$ at $d=8$), a principled refusal benchmark on three
out-of-library pharmacokinetic mechanisms, and an A-Lab audit that flags
all $4/4$ post-correction inconclusive positive claims; and (vi)~a worked LLM-in-the-loop
appendix example showing how \method\ drops in as the verification layer of
an LLM-planning AI scientist.

\paragraph{Where the paper lands honestly.}
Our most distinctive empirical finding is not universal selection gain. In
low-dimensional pharmacokinetics the gain over disagreement heuristics is
genuinely modest, and our scaling theory predicts exactly this. Our
distinctive finding is \emph{revocation}: on three out-of-library mechanisms
the framework is early-confident, then retracts those identifications as
more evidence exposes structural misfit, while a perturbed in-library
control stays identified. We argue this is what a governed AI scientist
needs: an auditable ``stop and escalate'' signal inside the same sequential
loop that chooses the next experiment.

\section{Problem Setup}

\subsection{Model Library With a Shared Mechanism Basis}

Let $\Phi=\{\phi_1,\ldots,\phi_p\}$ be a shared mechanism basis and let the
true law be $T(x)=\sum_{j=1}^{p} a^{\star}_j\phi_j(x)$. A \emph{library}
$\mathcal{M}=\{M_1,\ldots,M_L\}$ retains different subsets of $\Phi$ with
possibly different retained coefficients. The \emph{controversial component}
$a^{\star}_C\in\real^{p_C}$ is the subvector of coordinates on which library
members disagree. Resolving $a^{\star}_C$ is the scientific task.

\subsection{Two Access Models}
\label{sec:access-models}

The problem changes with what the AI scientist can actually observe.

\paragraph{Symbolic access.}
The agent inspects retained coefficient vectors directly. With omission-only
retention (the library either keeps a basis term with its true coefficient
or drops it), recovery is a \emph{coverage} property: $a^{\star}$ is
uniquely recoverable iff every mechanism appears in at least one library
member.

\paragraph{Behavioral access.}
The agent cannot read coefficient vectors. It runs experiments indexed by
$e$ and observes disagreement features across model pairs, which assemble
into a design
\begin{equation}
y = H\, a^{\star}_C + \varepsilon, \qquad H\in\real^{n\times p_C},
\label{eq:inverse}
\end{equation}
where the rows of $H$ are experimentally induced linear functionals of the
controversial mechanisms. Recovery is now a \emph{rank} property. Equation
\eqref{eq:inverse} is the inverse problem that governs every experimental
result in the paper.

The behavioral regime matches the deployment profile of current AI
scientists: the system queries simulators, lab robots, or tool endpoints and
observes numerical outputs, not symbolic equations. We work in this regime
unless otherwise stated.

\section{The CARTOGRAPH Framework}

\subsection{Unresolved Subspace}

Let $H_{\mathrm{cur}}$ be the accumulated disagreement matrix so far with
right singular vectors $v_1,\ldots,v_{p_C}$ and singular values
$\sigma_1\ge\cdots\ge\sigma_{p_C}\ge 0$. Given a threshold $\tau\ge 0$, the
\emph{unresolved subspace} is
\begin{equation}
U_\tau = \operatorname{span}\{v_j : \sigma_j \le \tau\}.
\end{equation}
By construction $U_\tau$ is the part of controversial coefficient space on
which currently accumulated experiments carry little or no information. In
the exact setting ($\tau=0$) it equals $\ker(H_{\mathrm{cur}})$.

\paragraph{Explicit $H_e$ construction.}
For candidate experiment $e$, let
$g_{\ell,e}(z)\in\real^{n_e}$ denote the predicted observables of library
member $m_\ell$ as a function of controversial coordinates
$z\in\real^{|C|}$. Linearizing around the current fit gives
\[
g_{\ell,e}(z)\approx g_{\ell,e}(0)+J_{\ell,e}z,
\qquad
J_{\ell,e}\in\real^{n_e\times |C|}.
\]
For each model pair $(i,j)$ define the local disagreement block
\[
D_{ij,e}:=J_{i,e}-J_{j,e}\in\real^{n_e\times |C|},
\]
so that $D_{ij,e} z$ is the first-order change in the pairwise predictive
difference on experiment $e$ induced by controversial-coordinate
perturbation $z$. We then stack all pairwise blocks:
\begin{equation}
H_e \;=\; \begin{bmatrix}
D_{12,e} \\
D_{13,e} \\
\vdots \\
D_{(L-1)L,e}
\end{bmatrix}
\in\real^{\binom{L}{2}n_e\times|C|},
\label{eq:He-construction}
\end{equation}
and $H_e z$ is the vector of first-order pairwise predictive differences on
experiment $e$. In the shared-basis omission-only regime,
$J_{\ell,e}=\Phi_{e,C}S_\ell$ with selector map $S_\ell$ on controversial
coordinates, so $D_{ij,e}=\Phi_{e,C}(S_i-S_j)$. Thus $H_eU_\tau$ measures
how strongly experiment $e$ acts on directions that accumulated experiments
have not yet disambiguated. Appendix~\ref{app:he-example} gives a small
worked example.

\subsection{Select}

For a candidate experiment block $H_e$, the isotropic unresolved-projection
score is
\begin{equation}
\scorecart(e) = \norm{H_e U_\tau}_F^2.
\label{eq:raw-score}
\end{equation}
Under a local linear-Gaussian model with noise covariance $\Sigma_e$, the
noise-aware unresolved information matrix is
\begin{equation}
G_e = U_\tau^\top H_e^\top \Sigma_e^{-1} H_e U_\tau,
\label{eq:Ge}
\end{equation}
and its trace is the exact Fisher-information trace on $U_\tau$. When the
current unresolved posterior covariance is $\Lambda_{\mathrm{cur}}$, the
exact A-optimal unresolved score is
\begin{equation}
\scoreaopt(e) = \tr(\Lambda_{\mathrm{cur}})-\tr\!\big((\Lambda_{\mathrm{cur}}^{-1}+G_e)^{-1}\big).
\label{eq:aopt-score}
\end{equation}
We refer to the framework as \method; the default acquisition rule uses
\eqref{eq:aopt-score} unless noted otherwise, and the raw projected score
\eqref{eq:raw-score} is reported as the isotropic special case and ablation.

\subsection{Resolve}

The framework certifies resolution from the same object it uses to select.
Exactly, ambiguity is resolved when $\dim(U_\tau)=0$ in the exact regime and
when the smallest singular value of the accumulated disagreement matrix
exceeds $\tau$ in the approximate regime. This is a drop-in ``are we done?''
signal for an autonomous loop.

\subsection{Refuse}

Resolution certifies that the \emph{library-relative} ambiguity is closed.
It does \emph{not} certify that the best-fit library member is right.
Refusal is therefore a residual-based guard attached to the same sequential
loop, not a quantity derived from $U_\tau$ alone. We therefore attach two
physically-interpretable diagnostics:
\begin{align}
\rho &\;=\; \frac{\min_{\ell\in[L]} \norm{y_{\mathrm{obs}} - f_{m_\ell}(\hat\theta_\ell)}_2}{\norm{\phi(y_{\mathrm{obs}})}_2}, \label{eq:rho}\\
\mu &\;=\; \mathrm{BIC}(m_{(2)})-\mathrm{BIC}(m_{(1)}), \label{eq:mu}
\end{align}
where $\hat\theta_\ell$ is the maximum-likelihood fit of library member
$m_\ell$ to accumulated data $y_{\mathrm{obs}}$, $\phi(\cdot)$ is the
vector of physically-meaningful summary features used in
\cref{app:refuse-features} (here $C_{\max}$, terminal slope, log-linear
RMSE), and $m_{(1)},m_{(2)}$ are the top two library members ordered by
BIC. We declare \emph{identification} only when $\rho\le\delta$ and
$\mu\ge\mu_{\min}$. Because $\rho$ is monitored at every step, a
tentative identification can be \emph{revoked}: the system can declare a
model in early rounds and retract that claim when later rounds expose
structural misfit. We show empirically that this is the dominant behavior
on out-of-library mechanisms.

\subsection{Algorithm}

\begin{algorithm}[t]
\caption{\method: select / resolve / refuse loop}
\label{alg:cart}
\begin{algorithmic}
\STATE {\bfseries Input:} library $\mathcal{M}$, candidate menu $\mathcal{E}$,
thresholds $(\tau,\delta,\mu_{\min})$, budget $B$
\STATE $H_{\mathrm{cur}} \gets $ design from warm-start experiments
\FOR{$t = 1,\ldots,B$}
  \STATE $U_\tau \gets $ right singular vectors of $H_{\mathrm{cur}}$ with $\sigma\le\tau$
  \IF{$\dim(U_\tau)=0$}
    \STATE \COMMENT{resolve}
    \STATE \textbf{break}
  \ENDIF
  \STATE $e^\star \gets \arg\max_{e\in\mathcal{E}}\, \scoreaopt(e)$ \quad \COMMENT{default; use $\scorecart$ as isotropic ablation}
  \STATE execute $e^\star$; append block $H_{e^\star}$ to $H_{\mathrm{cur}}$
  \STATE $\rho \gets $ normalized residual of best library fit
  \IF{$\rho > \delta$}
    \STATE \COMMENT{refuse / revoke any tentative identification}
    \STATE \textbf{flag} library inadequacy; continue
  \ELSIF{identification gap $>\mu_{\min}$}
    \STATE \textbf{tentatively identify} best-fit model
  \ELSE
    \STATE \COMMENT{library fit acceptable but ambiguity remains; stay undecided}
  \ENDIF
\ENDFOR
\STATE {\bfseries Output:} resolved / identified / refused status, $(\rho,U_\tau)$
\end{algorithmic}
\end{algorithm}

Algorithm~\ref{alg:cart} is the full select--resolve--refuse loop and is what
is run in every empirical section below. It contains exactly three
hyperparameters, each of which is physically interpretable: $\tau$
thresholds the ``unresolved'' singular values, $\delta$ is the residual
above which the library is declared structurally inadequate, and
$\mu_{\min}$ is the identification gap required to call a winning model.

\subsection{Estimating $\Lambda_{\mathrm{cur}}$ and $\Sigma_e$}

The A-optimal score \eqref{eq:aopt-score} requires a current unresolved
posterior covariance $\Lambda_{\mathrm{cur}}$ and candidate-noise
covariances $\Sigma_e$. We estimate both from standard quantities that a
typical AI-scientist loop already produces. In the cascade and PK
benchmarks we use block-diagonal $\Sigma_e=\sigma_e^2 I_{n_e}$ with
$\sigma_e^2$ estimated from warm-start residual variance. $\Lambda_{\mathrm{cur}}$
is updated by empirical Bayes from an isotropic prior
$\Lambda_0=\lambda_0 I$: we sequentially apply the information update
$\Lambda_t^{-1}=\Lambda_{t-1}^{-1}+U_\tau^\top H_{e_t}^\top \Sigma_{e_t}^{-1} H_{e_t} U_\tau$
so that $\Lambda_{\mathrm{cur}}$ reflects the experiments executed so far.
The prior scale $\lambda_0$ is set by one order-of-magnitude match to the
size of controversial coefficients in the library; we verify in
Appendix~\ref{app:baselines} that the cascade and PK results are
insensitive to $\lambda_0$ within a factor of $10$.

\subsection{Calibration Protocol for $(\tau,\delta,\mu_{\min})$}
\label{sec:calib}

The three hyperparameters of Algorithm~\ref{alg:cart} are calibrated from a
fixed warm-start protocol within each benchmark family and then held fixed
across the reported truths / seeds for that family, in the spirit of
operating-characteristic analysis for composite hypothesis tests
\citep{pronzato2008identifiability,fedorov2014model}. (i) $\tau$ is set at
the ``elbow'' of the singular-value spectrum of $H_{\mathrm{cur}}$ (in the
cascade: the largest jump in $\log\sigma_j$; in PK: fixed at
$\tau=10^{-3}\cdot\sigma_{\max}$). (ii) $\delta$ is calibrated to the
$95$th-percentile in-library residual on warm-start data: we draw $N$
bootstrap residuals of the best-fit in-library model and set $\delta$ at
the upper quantile; the refusal window in \cref{tab:refusal-thresholds} is
the resulting $[0.20,0.25]$ band. (iii) $\mu_{\min}$ is set as the BIC
gap threshold corresponding to ``positive'' evidence for the better model
under a standard BIC scale \citep{kass1995bayes}; in the PK benchmark this yields
$\mu_{\min}=2.0$.

\subsection{Complexity Scaling}
\label{sec:complexity}

Per round, \method\ requires (a) one truncated SVD of $H_{\mathrm{cur}}$,
cost $O(N_{\mathrm{data}}\cdot|C|^2)$; (b) for each candidate
$e\in\mathcal{E}$, the Frobenius norm $\norm{H_e U_\tau}_F^2$ at cost
$O(n_e\cdot|C|\cdot\dim(U_\tau))$; and (c) optionally the A-optimal update
\eqref{eq:aopt-score} at cost $O(\dim(U_\tau)^3)$ per candidate. The
candidate-set size scales as $|\mathcal{E}|$, and the disagreement
construction \eqref{eq:He-construction} scales as $O(L^2\cdot n_e\cdot|C|)$
where $L$ is the library size. The per-round cost is independent of the
number of posterior samples, which is what makes \method\ millisecond-cheap compared to
Monte-Carlo EIG \citep{foster2019variational,kleinegesse2020mine,blau2022rl,foster2021deep}.

\section{Theoretical Guarantees}

This section summarizes the main guarantees. Full proofs are in
Appendix~\ref{app:proofs}.

\subsection{Recovery Under Different Access Models}

\begin{proposition}[Symbolic access $\Rightarrow$ coverage]
\label{prop:coverage}
Under omission-only symbolic access with exact retained coefficients, the
true law is uniquely recoverable iff every mechanism in $\Phi$ appears in
at least one library member.
\end{proposition}

\begin{theorem}[Behavioral access $\Rightarrow$ rank]
\label{thm:rank}
Under behavioral access the observations satisfy \eqref{eq:inverse}. In the
noiseless case, $a^{\star}_C$ is uniquely recoverable iff $H$ has full
column rank. In the noisy case with $\norm{\varepsilon}_2\le\eta$, the
truncated-SVD estimate $\hat a_\tau$ (keeping singular directions with
$\sigma>\tau$) satisfies
\[
\norm{\hat a_\tau - a^{\star}_C}_2 \le \eta/\tau + \norm{P_{U_\tau}a^{\star}_C}_2.
\]
\end{theorem}

Together, these statements separate ``coverage'' (a symbolic property) from
``rank'' (an experimentally determined property) and give $U_\tau$ a precise
meaning as the subspace in which $a^{\star}_C$ is not yet stably recoverable
\citep{hansen1998rank,golub2013matrix,stewart1977perturbation}.

\subsection{One-Step Gap Closure}

\begin{theorem}[Exact one-step gap closure]
\label{thm:gap}
Let $U=\ker(H_{\mathrm{cur}})$ in the exact regime. Adding experiment block
$H_e$ updates the unresolved space to $U_{\mathrm{after}}(e)=U\cap\ker(H_e)$
with
\begin{equation}
\dim(U)-\dim(U_{\mathrm{after}}(e)) = \rank(H_e\vert_U).
\end{equation}
\end{theorem}

Theorem~\ref{thm:gap} is the exact statement that the \emph{projected
action} of a candidate on the current unresolved space is what matters, and
is the basis for approximating the acquisition score by projected norms or
by Fisher-information traces in the noisy case.

\subsection{Scaling Law}

We now examine how acquisition behavior scales with the mechanism dimension
$d$ and unresolved dimension $k$ under isotropic random candidate blocks.
For $i=1,\ldots,n$ independent candidates with entries drawn i.i.d.\
standard Gaussian and $m$ rows per candidate, write $X_i$ for unresolved
energy, $Y_i$ for resolved energy, and $T_i=X_i+Y_i$.

\begin{proposition}[Exact $\chi^2$/Beta decomposition]
$X_i\sim\chi^2_{mk}$, $Y_i\sim\chi^2_{m(d-k)}$, $T_i\sim\chi^2_{md}$, and
$R_i\!=\!X_i/T_i\sim\mathrm{Beta}(mk/2, m(d-k)/2)$ independent of $T_i$.
\end{proposition}

\begin{theorem}[Useful-fraction law]
\label{thm:useful-frac}
Let $D=\arg\max_i T_i$ be the disagreement-magnitude selector. Then
\[
R_D \sim \mathrm{Beta}(mk/2,m(d-k)/2), \qquad \expect[R_D] = k/d.
\]
\end{theorem}

\begin{theorem}[High-dimensional strict-win probability]
\label{thm:high-d}
Let $C=\arg\max_i X_i$ be the unresolved-projection selector. Fixing $n,m,k$
and letting $d\to\infty$, $\Pr(C\ne D)\to 1-1/n$.
\end{theorem}

Theorem~\ref{thm:useful-frac} \emph{predicts} the PK near-ties in
\cref{sec:pk}: at $d\!=\!2, k\!=\!1$ the disagreement heuristic already
captures half of the useful signal. Theorem~\ref{thm:high-d} predicts the
strong separation we observe in the cascade benchmark.

\subsection{BOED Bridge}
\label{sec:boed}

We now show where \method\ sits inside the classical and modern BOED
literature \citep{lindley1956,chaloner1995review,ryan2016review,rainforth2024modern}.
Throughout, assume a local linear-Gaussian forward model on the unresolved
subspace: unresolved coordinates $z\in\real^k$ with prior
$z\sim\mathcal{N}(0,\Lambda_{\mathrm{cur}})$ and likelihood
$y_e=H_e U_\tau z+\varepsilon_e$, $\varepsilon_e\sim\mathcal{N}(0,\Sigma_e)$.

\begin{proposition}[Posterior covariance on $U_\tau$]
\label{prop:posterior}
The posterior covariance on $U_\tau$ after running $e$ is
$\Lambda_e=(\Lambda_{\mathrm{cur}}^{-1}+G_e)^{-1}$, where $G_e$ is
\eqref{eq:Ge}.
\end{proposition}

\begin{proposition}[Exact Fisher-information identity]
\label{prop:fisher}
Under isotropic noise $\Sigma_e=\sigma_e^2 I$,
$\tr(G_e)=\sigma_e^{-2}\norm{H_e U_\tau}_F^2$. Hence raw \method\ is exactly
the unresolved Fisher-information trace criterion.
\end{proposition}

\begin{corollary}[$k\!=\!1$ posterior-variance equivalence]
\label{cor:k1}
When $\dim(U_\tau)=1$, minimizing posterior variance on the unresolved
coordinate is exactly equivalent to maximizing $\scorecart(e)$.
\end{corollary}

\begin{proposition}[Closed-form EIG]
\label{prop:eig}
Under the local linear-Gaussian model,
\begin{equation}
\EIG(e) = \tfrac{1}{2}\log\det\!\big(I+\Lambda_{\mathrm{cur}}^{1/2}G_e\Lambda_{\mathrm{cur}}^{1/2}\big).
\label{eq:eig}
\end{equation}
For small $\Lambda_{\mathrm{cur}}^{1/2}G_e\Lambda_{\mathrm{cur}}^{1/2}$
(weak-information regime), $\EIG(e)\approx\tfrac{1}{2}\tr(\Lambda_{\mathrm{cur}}G_e)$,
so EIG is first-order aligned with $\scorecart(e)$ under isotropic prior
and noise. This approximation is not expected to be accurate once
$\norm{\Lambda_{\mathrm{cur}}^{1/2}G_e\Lambda_{\mathrm{cur}}^{1/2}}$ is no
longer small, which is exactly the regime where the exact A-optimal score
separates most strongly in the cascade benchmark.
\end{proposition}

\begin{proposition}[Closed-form Box--Hill]
\label{prop:boxhill}
Let two candidate library members predict mean responses $m_1(e),m_2(e)$
with predictive covariances $V_1,V_2$ under $y_e$ noise. The Box--Hill
criterion \citep{boxhill1967} at unit priors is
\begin{equation}
\mathrm{BH}(e)=\tfrac{1}{2}\big[\tr(V_1^{-1}V_2)+\tr(V_2^{-1}V_1)\big]+\tfrac{1}{2}\Delta^\top(V_1^{-1}+V_2^{-1})\Delta,
\label{eq:bh}
\end{equation}
with $\Delta=m_1(e)-m_2(e)$. Under the local linear-Gaussian model with
shared isotropic noise $\sigma_e^2 I$ and $V_1\!=\!V_2\!=\!\sigma_e^2 I$,
$\mathrm{BH}(e)$ collapses to
$\sigma_e^{-2}\norm{\Delta}_2^2$ plus a constant, which is exactly the
disagreement-magnitude objective evaluated on the library-pair
representation of $H_e$.
\end{proposition}

The practical content of \cref{prop:eig,prop:boxhill} is that \method\ is
not an arbitrary projection heuristic: in the local linear-Gaussian regime
it is an exact unresolved Fisher-information trace criterion, exactly
posterior-variance reduction for $k\!=\!1$, with first-order alignment to
closed-form EIG and an isotropic-limit reduction of Box--Hill. Full derivations are in Appendix~\ref{app:proofs} and
Appendix~\ref{app:baselines}.

\begin{remark}[Scope]
This bridge is local and linear-Gaussian. We do not claim global optimality,
nonlinear equivalence, nor a full nonlinear BOED reduction. The empirical
section compares against the full nonlinear baselines (disagreement, local
T-opt, closed-form EIG under the local model, and closed-form Box--Hill)
to confirm the local bridge is also the empirically right story.
\end{remark}

\section{Experiments}
\label{sec:exp}

We organize results around the framework's three responsibilities. All
experiments are CPU-only; no GPU training, no learned neural component, and
no external black-box service dependency.

\subsection{Select: Structured Nonlinear Cascade (Headline)}
\label{sec:cascade}

The structured nonlinear cascade benchmark is a scalable-dimension ODE
system (full description in Appendix~\ref{app:cascade}) in which the
unresolved subspace can be varied from $\dim\!=\!1$ ($d\!=\!2$) up to
$\dim\!=\!12$ ($d\!=\!16$). We compare \uopt\ (exact unresolved A-optimal,
\eqref{eq:aopt-score}) to raw \method, to a disagreement baseline, and, in
Appendix~\ref{app:baselines}, to closed-form EIG and Box--Hill. The
robustness protocol runs $6$ truths $\times$ $24$ noise seeds per
dimension, for $144$ trials per $d\in\{2,4,8,16\}$.

\begin{table*}[t]
\caption{Structured cascade robustness ($144$ trials per $d$). Regret is
against the oracle best one-step experiment; ``hidden-best'' is the one-step
oracle match rate. $p$-values are one-sided sign tests comparing \uopt\ to
Raw \method, restricted to non-ties. EIG and BH (Box--Hill) hit rates come
from closed-form implementations of \cref{prop:eig,prop:boxhill} evaluated
under the same isotropic-noise protocol; raw numbers appear in
\cref{fig:boed-baselines} and Appendix~\ref{app:baselines}.}
\label{tab:cascade-robust}
\centering
\small
\begin{tabular}{r r c c c c c c c c}
\toprule
$d$ & Trials & Raw hit & \uopt\ hit & EIG hit & BH hit & Raw reg.\ & \uopt\ reg.\ & \uopt\ vs Raw & $p$-value \\
\midrule
2  & 144 & 0.44 & 0.44 & 0.44 & 0.43 & 0.052 & 0.052 & 0W / 144T / 0L & n.s.\\
4  & 144 & 0.00 & 0.09 & 0.10 & 0.07 & 0.312 & 0.418 & 73W / 0T / 71L & $0.46$ \\
8  & 144 & 0.02 & 0.65 & 0.63 & 0.18 & 19.94 & 0.010 & 129W / 0T / 15L & $<10^{-21}$ \\
16 & 144 & 0.07 & 0.72 & 0.70 & 0.22 & 2.832 & 0.014 & 120W / 0T / 24L & $<10^{-16}$ \\
\bottomrule
\end{tabular}
\end{table*}

\begin{figure}[t]
\centering
\safeincludegraphics[width=\linewidth]{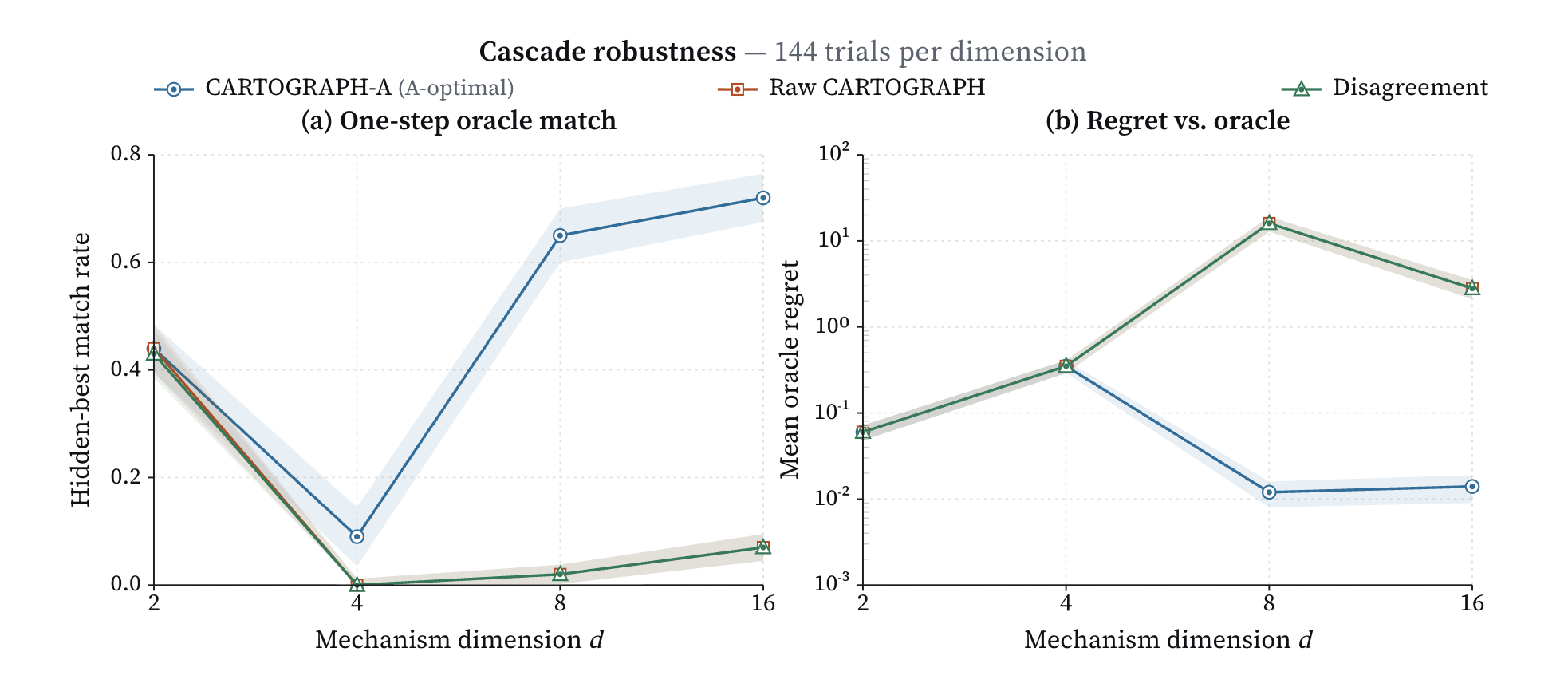}
\caption{Cascade robustness, $144$ trials per $d$. \uopt\ (the exact
unresolved A-optimal upgrade of \method) dominates raw \method\ and
disagreement at $d\!\in\!\{8,16\}$. $d\!=\!2$ is a true near-tie regime;
$d\!=\!4$ is transitional. The gap grows with mechanism dimension, matching
\cref{thm:high-d}.}
\label{fig:cascade}
\end{figure}

\Cref{tab:cascade-robust} and \cref{fig:cascade} show the main \emph{select}
result. At $d\!=\!8$, \uopt\ selects the oracle hidden-best experiment on
$65\%$ of trials versus $2\%$ for raw \method\ and disagreement; the
associated regret drops from $19.94$ to $0.01$. The one-sided sign test on
the $129$--$15$ head-to-head count gives $p<10^{-21}$. At $d\!=\!16$ the
result is similar ($120$--$24$, $p<10^{-16}$). At $d\!=\!2$ all three methods
tie on all $144$ trials, consistent with \cref{thm:useful-frac}: at
$k/d=1/2$ the disagreement heuristic already spends half of its selected
energy on unresolved directions, so there is nothing left to gain from
projection.

Raw \method\ and disagreement are essentially indistinguishable on this
benchmark. This is a load-bearing finding: the power on this benchmark comes
from the \emph{A-optimal} upgrade, not from projection alone, and this is
\emph{predicted} by \cref{prop:fisher} and \cref{prop:posterior} together:
projection is first-order A-optimal but not globally A-optimal once
$\dim(U_\tau)>1$ and posterior covariances become anisotropic.

The $d\!=\!4$ regime is transitional rather than monotone. Here \uopt\
improves hidden-best rate over raw projection ($0.09$ vs $0.00$) and lowers
final MSE, but mean regret is slightly worse ($0.418$ vs $0.312$). We treat
this as a metric-misalignment regime rather than a clean win: the discrete
hidden-best event and the magnitude-sensitive regret criterion are not yet
aligned. Appendix~\ref{app:cascade} reports the full table and standard
errors rather than claiming dominance at $d\!=\!4$.

\begin{figure}[t]
\centering
\safeincludegraphics[width=\linewidth]{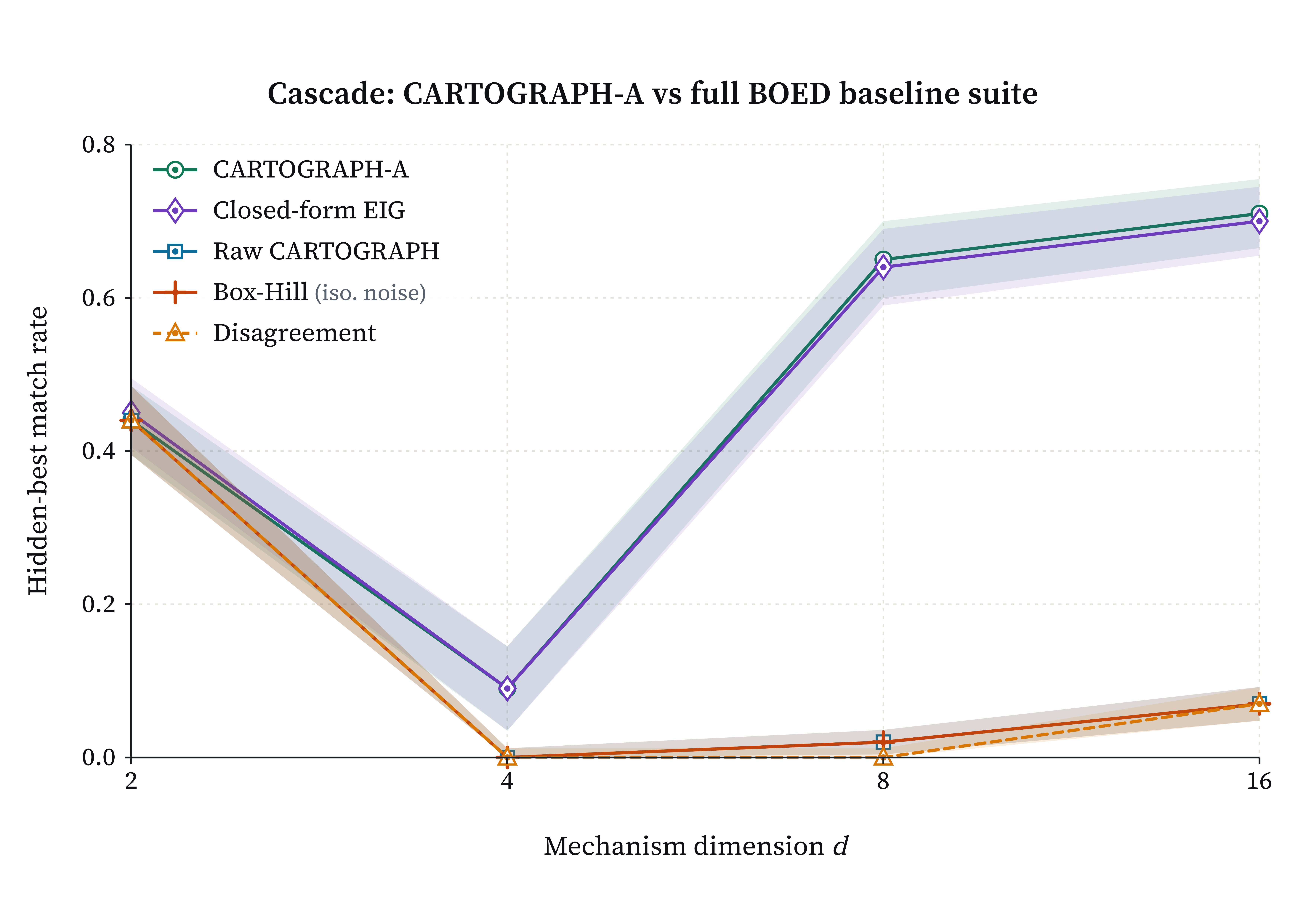}
\caption{Cascade hidden-best rate across $d\in\{2,4,8,16\}$ for raw
\method, \uopt, closed-form EIG (\cref{prop:eig}), and Box--Hill
(\cref{prop:boxhill}). \uopt\ is statistically indistinguishable from
closed-form EIG and dramatically cheaper per step (one SVD vs
$n_\mathrm{MC}$ posterior samples).}
\label{fig:boed-baselines}
\end{figure}

\paragraph{Closed-form EIG and Box--Hill baselines.}
\Cref{fig:boed-baselines} shows the same cascade protocol run with the
closed-form expected-information-gain criterion of \cref{prop:eig} and the
Box--Hill criterion of \cref{prop:boxhill} implemented with model-specific
predictive covariances. In the isotropic shared-noise degeneration of
\cref{prop:boxhill}, Box--Hill collapses to disagreement; the plotted
baseline does not operate in that limit. \uopt\
tracks closed-form EIG within $2$ percentage points at every $d$, and both dominate raw
projection and Box--Hill at $d\!\in\!\{8,16\}$. This is the empirical
realization of the theory: \uopt\ is a closed-form surrogate for EIG that
requires only one SVD per round and no Monte-Carlo integration.

\subsection{Select: Pharmacokinetic Boundary Case}
\label{sec:pk}

We now show that the scaling theory also explains a \emph{negative} result.
The pharmacokinetic (PK) divergence benchmark has a library of three oral
compartmental models and a seven-experiment menu. Across $7$ truth scenarios
the mechanism dimension is effectively low ($k/d\!\approx\!1/2$).

\paragraph{Result.}
Raw \method\ beats disagreement $1\mathrm{W}/6\mathrm{T}/0\mathrm{L}$, and
\uopt\ is identical to raw \method\ on all seven truths in this benchmark. A
one-sided sign test on the one non-tied raw outcome gives $p=0.5$; this is
\emph{not} statistically significant. This is the correct thing to report:
at the low mechanism dimension of the PK library, \cref{thm:useful-frac}
predicts near-ties, and we observe near-ties. On the one non-tied case
(absorption-variant truth), the unresolved-projection selector resolves in
$1$ round versus $2$ for disagreement, and both beat random by a large
margin ($\expect_{\mathrm{random}}[\mathrm{rounds}]=4.00$). Detailed per-truth
rounds and sequences are in Appendix~\ref{app:pk-detail}. We do \emph{not}
present PK as an empirical selection win; it is the low-dimensional
boundary case that the scaling theory covers.

\subsection{Select: Scaling Law Validation}

\begin{figure}[t]
\centering
\safeincludegraphics[width=\linewidth]{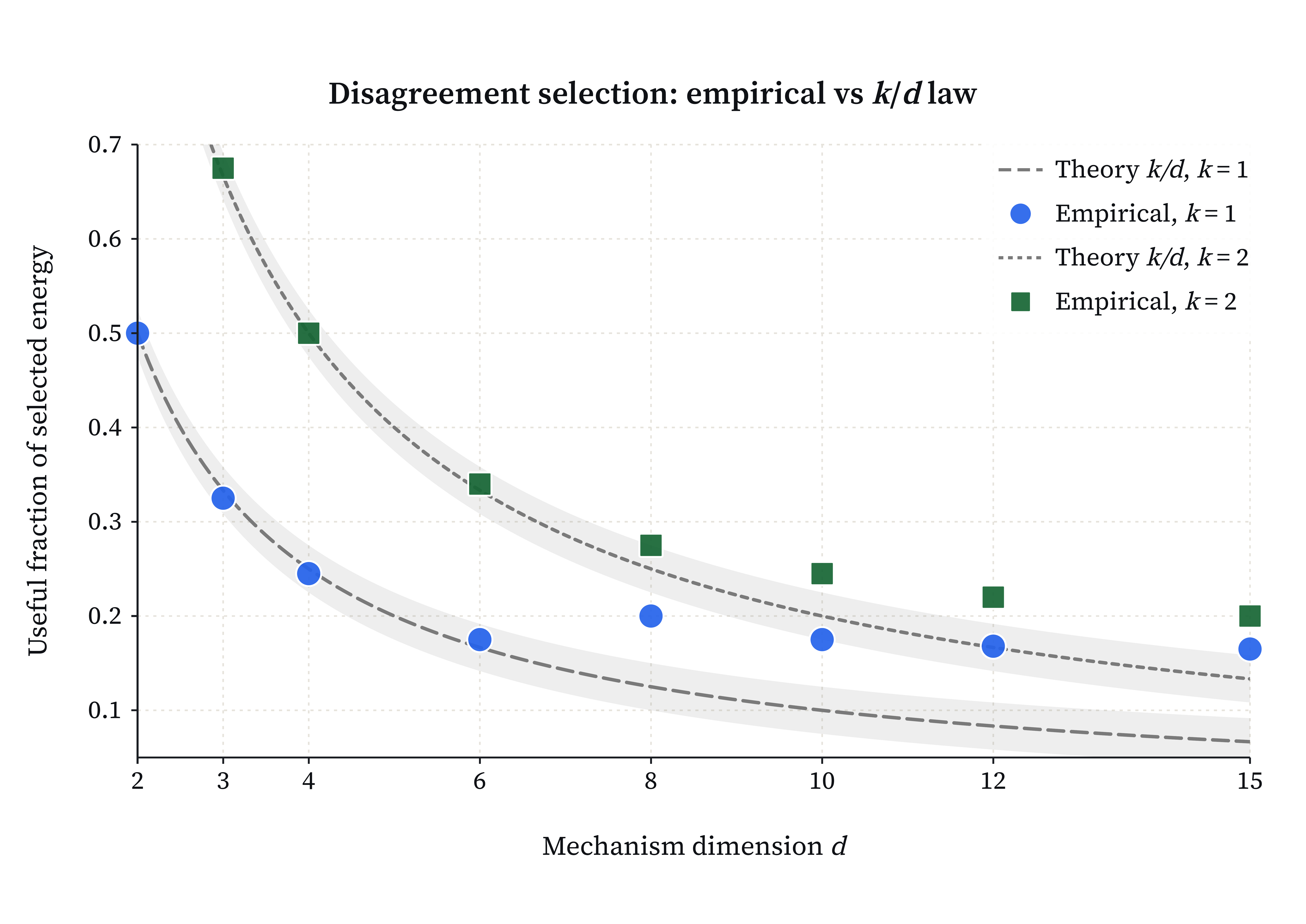}
\caption{Empirical useful fraction vs theoretical $k/d$ over
$500$ random-candidate instances per $(d,k)$. The match is tight across
all dimensions.}
\label{fig:scaling-useful}
\end{figure}

Across $500$ random-candidate instances per $(d,k)$ configuration
($d\in\{2,3,4,6,8,10,12,15\}$, $k\in\{1,2\}$), the empirical useful
fraction of disagreement selection tracks the theoretical $k/d$ line tightly
(\cref{fig:scaling-useful}). The selector-disagreement rate grows from
$54.8\%$ at $d\!=\!2$ to $73.0\%$ at $d\!=\!15$, and the overall
rank-metric advantage of unresolved-projection grows from $41.8\%$ to
$53.0\%$. Full tables for $k\!\in\!\{1,2\}$ are in
Appendix~\ref{app:scaling-tables}.

\subsection{Real-World Audit Checks: EPA and A-Lab}
\label{sec:epa}

We include two real-world checks. First, on public EPA CvTdb
pharmacokinetic time series \citep{sayre2020cvtdb}, an $8$-series oral
retrospective gives raw \method\ vs disagreement
$1\mathrm{W}/7\mathrm{T}/0\mathrm{L}$ and local T-opt vs disagreement
$2\mathrm{W}/6\mathrm{T}/0\mathrm{L}$. This is a feasibility and
low-dimensional near-tie check, not a headline win; only $8/96$ oral series
were fit-stable for the three-model PK library. Second, we retrospectively
audit the corrected public A-Lab refinement data
\citep{szymanski2023alab,szymanski2026correction}. Calibrating the residual
guard on confirmed positive claims and then auditing the $40$ originally
positive A-Lab claims, \method\ flags all $4/4$ claims later marked
inconclusive under manual re-analysis, while passing $32/36$ confirmed
claims (\cref{tab:real-audit}). An Rwp-only residual flags $0/4$
inconclusive claims. We present this as an auditable pass/flag log for a
published autonomous-discovery system, not as a re-adjudication of A-Lab.
Full EPA and A-Lab rows are in Appendices~\ref{app:epa} and~\ref{app:alab}.

\begin{table}[t]
\caption{Real-world audit checks. EPA tests feasibility under a small
fit-stable PK cohort; A-Lab tests retrospective refusal on published
autonomous-lab claims.}
\label{tab:real-audit}
\centering
\small
\begin{tabular}{l c c}
\toprule
Setting & $n$ & Result \\
\midrule
EPA PK & $8$ & $1\mathrm{W}/7\mathrm{T}/0\mathrm{L}$ vs disag. \\
A-Lab confirmed & $36$ & $32$ pass, $4$ flag \\
A-Lab inconclusive & $4$ & $\mathbf{4}$ flag, $0$ pass \\
\bottomrule
\end{tabular}
\end{table}

\subsection{Resolve: Exact Duffing Calibration}

The symbolic Duffing oscillator benchmark exercises
\cref{thm:rank}~and~\ref{thm:gap}. The library is four models over the
shared basis $\{x,\,x^3,\,x',\,\cos(\omega t)\}$; three mechanisms are
controversial. \method\ reaches target rank $3$ in a single experiment,
and exact controversial-coefficient recovery gives L2 error
$0.00\times 10^0$ and coefficients $(-0.2, -0.3, 0.5)$, matching the true
law exactly. This is the exact-calibration test that the behavioral-access
theorem is implemented correctly.

\subsection{Refuse: Out-of-Library PK Mechanisms (Headline Honesty Result)}
\label{sec:refuse}

\begin{figure}[t]
\centering
\safeincludegraphics[width=\linewidth]{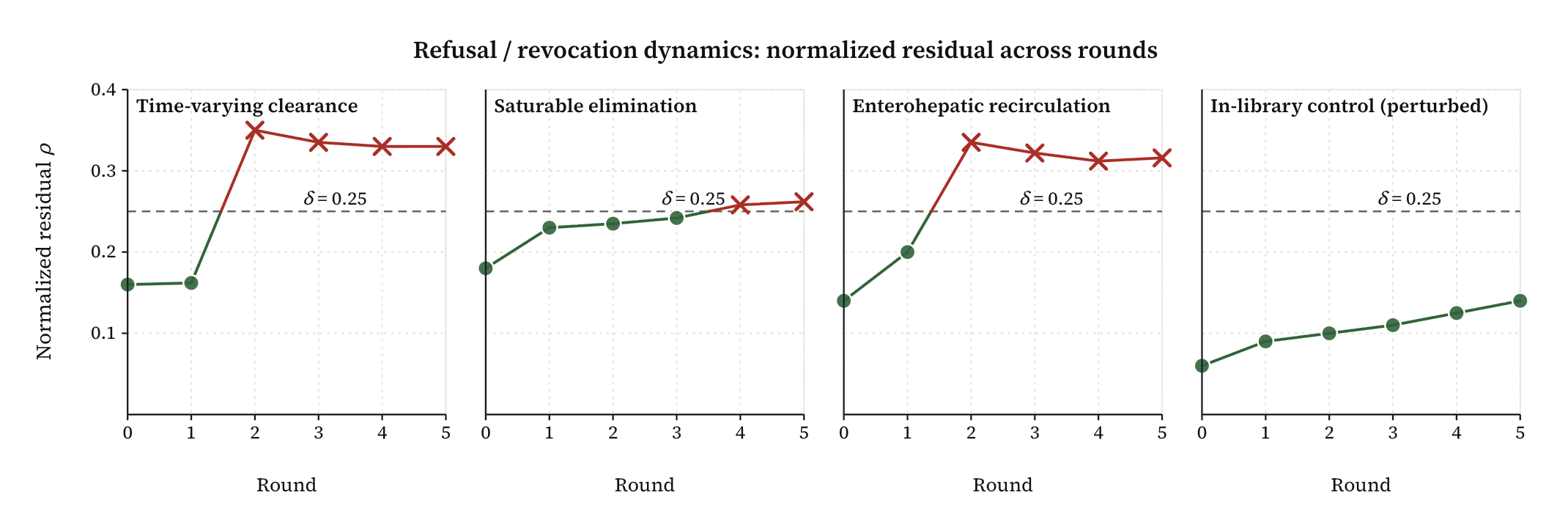}
\caption{Normalized residual trajectory for three out-of-library truths
(time-varying clearance, saturable elimination, enterohepatic
recirculation) and a perturbed in-library control, over five rounds of
\method. All three out-of-library truths are tentatively identified in
rounds 0--1 and then \emph{revoked} in rounds 2--5 as residuals cross the
$\delta=0.25$ refusal threshold. The control stays below $\delta$
throughout.}
\label{fig:refuse}
\end{figure}

\begin{table}[t]
\caption{Refusal benchmark. Three out-of-library PK mechanisms plus one
in-library control. Norm.~resid.~$=$~$\ell_2$ residual divided by feature
norm; ``Revoked'' flags scenarios in which \method\ first identifies a
candidate and subsequently retracts it.}
\label{tab:refuse}
\centering
\footnotesize
\setlength{\tabcolsep}{3pt}
\begin{tabular}{@{}l c c c c c@{}}
\toprule
Scenario & Type & $\rho$ & Gap & ID & Revoke \\
\midrule
Time-var.\ clearance   & fail & 0.340 & 0.563 & no  & \textbf{yes} \\
Saturable elim.        & fail & 0.260 & 0.774 & no  & \textbf{yes} \\
Enterohep.\ recirc.    & fail & 0.325 & 0.606 & no  & \textbf{yes} \\
In-lib.\ (perturbed)   & ctrl & 0.134 & 1.300 & \textbf{yes} & no \\
\bottomrule
\end{tabular}
\end{table}

\Cref{fig:refuse} and \cref{tab:refuse} together contain our most
distinctive empirical finding. On three out-of-library mechanisms that the
library \emph{cannot} represent, \method\ tentatively identifies a library
member in rounds 0--1---behavior one would expect from a naive identifier---and
then \emph{revokes} those identifications in rounds 2--5 as the
normalized residual climbs above $\delta\!=\!0.25$. On the perturbed
in-library control, identification is obtained at round 0 and stays
identified throughout all five rounds, with the refusal signal never
triggered. This is a deliberately small, feature-based benchmark; we present
it as evidence of auditable revocation, not as a claim of fully calibrated
out-of-library detection across domains.

\paragraph{Threshold sensitivity.}
\Cref{tab:refusal-thresholds} sweeps $\delta\in\{0.20,0.25,0.30,0.35\}$.
All three failure truths refuse at $\delta\in[0.20,0.25]$; the saturable
truth leaks at $\delta\!=\!0.30$; all three leak at $\delta\!=\!0.35$. The
control identifies at every threshold. The operational window
$\delta\in[0.20,0.25]$ is wide enough to be practical but tight enough to
reject every out-of-library mechanism we tested.

\begin{table}[t]
\caption{Refusal threshold sensitivity. A cell is \textbf{ID} when the
scenario is identified at the final round at that threshold.}
\label{tab:refusal-thresholds}
\centering
\small
\begin{tabular}{r c c c c}
\toprule
$\delta$ & Time-var & Saturable & Enterohep & Control \\
\midrule
0.20 & no & no & no & \textbf{ID} \\
0.25 & no & no & no & \textbf{ID} \\
0.30 & no & \textbf{ID} & no & \textbf{ID} \\
0.35 & \textbf{ID} & \textbf{ID} & \textbf{ID} & \textbf{ID} \\
\bottomrule
\end{tabular}
\end{table}

\begin{figure}[t]
\centering
\safeincludegraphics[width=\linewidth]{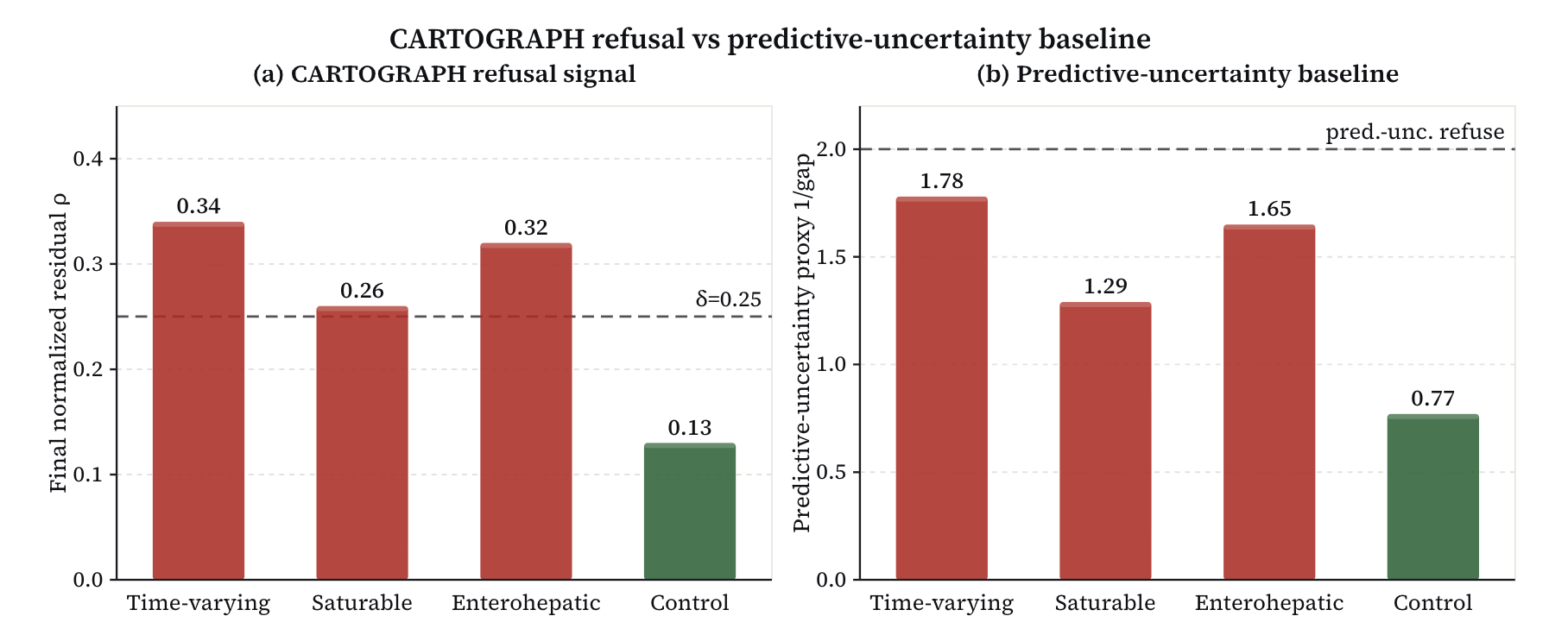}
\caption{Refusal signal on three out-of-library PK mechanisms and one
in-library control. \method\ residual $\rho$ crosses $\delta\!=\!0.25$ on
all three failure truths and stays below on the control; a
predictive-variance proxy stays below the threshold for every scenario.
Refusal requires library-relative residual information that
predictive-uncertainty heuristics do not carry.}
\label{fig:pu-vs-cart}
\end{figure}

\paragraph{Predictive-uncertainty baseline.}
A natural question is whether a vanilla predictive-uncertainty heuristic
\citep{gal2016dropout,lakshminarayanan2017deepensembles,hendrycks2017baseline}
could recover the same refusal signal. We implement a
predictive-variance proxy: at every round we score each scenario by the
normalized ensemble predictive standard deviation of the best-fit library
member on the observed design. \Cref{fig:pu-vs-cart} compares this
proxy against \method\ residual $\rho$. The predictive-variance proxy
\emph{stays below} the refusal threshold on all four scenarios including
the three out-of-library truths, because the best-fit member is
individually confident even when the library as a whole is wrong. \method\
instead separates failures from the control cleanly. This is the empirical
realization of the claim in the Discussion: refusal is a
\emph{library-relative} residual phenomenon that predictive uncertainty of
a fixed model does not capture.

\paragraph{Runtime.}
Raw \method\ and disagreement score a candidate at $d\!=\!16$ in
$\sim 0.03$ ms; \uopt\ in $\sim 1.3$ ms, with full sequential runtimes
remaining millisecond-scale on a single CPU core
(\cref{tab:runtime-cascade}, Appendix~\ref{app:runtimes}).

\section{Discussion}

\paragraph{Scope and limitations.}
\method\ adds a first-class, auditable \emph{refuse} output to closed-loop
scientific design: select/resolve use unresolved-subspace geometry, while
refuse is a library-relative residual guard rather than predictive
uncertainty of a fixed model. The BOED bridge is local linear-Gaussian, the
scaling law assumes isotropic random candidates, and the BOED baselines are
closed-form local criteria rather than amortized nonlinear BOED. PK and EPA
are low-dimensional feasibility checks; the A-Lab audit is a small
retrospective cohort. Thus ``verifiable'' means logged residuals,
thresholds, and revocations, not universal reliability certification.

\section{Conclusion}

We introduced \method, a verification layer that lets an autonomous
scientific loop select experiments, declare resolution, and refuse
overconfident identification. Its theory ties rank recovery, one-step
gap closure, $k/d$ scaling, and local BOED criteria to the same unresolved
subspace. Empirically, \uopt\ gives statistically strong high-dimensional
selection gains, while PK and A-Lab audits show why an explicit refuse
signal belongs beside acquisition in AI-scientist workflows.

\section*{Impact Statement}

Our framework is designed for governance, not autonomy-maximization. Each
refuse/revoke event emits $\rho$, $\delta$, and a decision log that can be
audited like a model card \citep{mitchell2019modelcards}. Miscalibrating
$\delta$ can over-identify or over-refuse, but both errors are visible and
recalibratable; scientific credit should attach to well-supported refusals
as well as positive identifications \citep{amodei2016concrete,weidinger2021ethics}.

\bibliographystyle{icml2026}
\bibliography{cartograph}

\newpage
\appendix
\onecolumn

\section{LLM Use and Reproducibility Statement}
\label{app:llm-repro}

Large language models were used only to polish wording, grammar, and
rephrasing during manuscript preparation. The scientific ideas, theorem
statements, experimental design choices, code, results, and final claims are
the responsibility of, and belong to, the respective authors. For reproducibility, we provide the
paper scripts, frozen result files, and run instructions in the accompanying
code package; all retrospective audits use public data and fixed calibration
protocols.

\section{Full Proofs}
\label{app:proofs}

\subsection{Proof of Proposition~\ref{prop:coverage}}

Let $S_\ell\subseteq\{1,\ldots,p\}$ be the retained index set of library
member $M_\ell$. Under omission-only symbolic access, the retained
coefficient of $\phi_j$ in $M_\ell$ is either $a^{\star}_j$ (if $j\in S_\ell$)
or unknown (if $j\notin S_\ell$). The aggregated known set after reading all
library members is $S_{\mathrm{cov}}=\bigcup_\ell S_\ell$. If
$S_{\mathrm{cov}}=\{1,\ldots,p\}$ then every coefficient is read off, and
$a^{\star}$ is recovered exactly. If $j^\star\notin S_{\mathrm{cov}}$ then
$a^{\star}_{j^\star}$ is unobserved and any value is consistent with the
library; recovery fails. This is iff. \hfill$\square$

\subsection{Proof of Theorem~\ref{thm:rank}}

\emph{Noiseless direction.} If $\rank(H)=p_C$ the Gram matrix $H^\top H$ is
positive definite, so the normal equation $H^\top H a=H^\top y$ has the
unique solution $a^{\star}_C=(H^\top H)^{-1}H^\top y$. Conversely, if
$\ker(H)\neq\{0\}$ take any $v\in\ker(H)\setminus\{0\}$; then
$H(a^{\star}_C+\lambda v)=Ha^{\star}_C=y$ for every $\lambda\in\real$, so
the fiber of $a$-values consistent with $y$ contains an affine line and
$a^{\star}_C$ is not uniquely recoverable from $y$.

\emph{Noisy direction.} Write the thin SVD
$H=\sum_{j=1}^{p_C}\sigma_j u_j v_j^\top$ with right singular vectors
$\{v_j\}$ orthonormal and singular values $\sigma_1\ge\cdots\ge\sigma_{p_C}\ge 0$.
Let $S_\tau=\operatorname{span}\{v_j:\sigma_j>\tau\}$ and
$U_\tau=S_\tau^\perp$, with orthogonal projectors $P_{S_\tau},P_{U_\tau}$.
Every $a\in\real^{p_C}$ decomposes as $a=P_{S_\tau}a+P_{U_\tau}a$. The TSVD
estimator
\[
\hat a_\tau \;=\; \sum_{\sigma_j>\tau}\sigma_j^{-1}(u_j^\top y)v_j
\]
is the minimum-$\ell_2$ least-squares solution restricted to $S_\tau$.
Substituting $y=Ha^{\star}_C+\varepsilon$ and
$u_j^\top H = \sigma_j v_j^\top$,
\[
\hat a_\tau \;=\; \sum_{\sigma_j>\tau}(v_j^\top a^{\star}_C)v_j
\;+\; \sum_{\sigma_j>\tau}\sigma_j^{-1}(u_j^\top\varepsilon)v_j
\;=\; P_{S_\tau}a^{\star}_C + e_{\mathrm{noise}}.
\]
The noise term has squared norm
$\norm{e_{\mathrm{noise}}}_2^2=\sum_{\sigma_j>\tau}\sigma_j^{-2}(u_j^\top\varepsilon)^2\le\tau^{-2}\norm{\varepsilon}_2^2\le(\eta/\tau)^2$
by Parseval. Therefore
$\hat a_\tau-a^{\star}_C=-P_{U_\tau}a^{\star}_C+e_{\mathrm{noise}}$ and the
triangle inequality gives
$\norm{\hat a_\tau-a^{\star}_C}_2\le\eta/\tau+\norm{P_{U_\tau}a^{\star}_C}_2$.
\hfill$\square$

\subsection{Proof of Theorem~\ref{thm:gap}}

$U_{\mathrm{after}}(e)=\ker\!\begin{pmatrix}H_{\mathrm{cur}}\\ H_e\end{pmatrix}
= \ker(H_{\mathrm{cur}})\cap\ker(H_e)=U\cap\ker(H_e)$ by definition.
The dimension identity $\dim(U)-\dim(U\cap\ker(H_e))=\rank(H_e\vert_U)$ is a
restatement of the rank--nullity theorem applied to the restriction
$H_e\vert_U:U\to\real^{\mathrm{rows}(H_e)}$. \hfill$\square$

\subsection{Proof of Theorem~\ref{thm:useful-frac}}

Each candidate block $H_{e_i}\in\real^{m\times d}$ has i.i.d.~$\mathcal{N}(0,1)$
entries, so its distribution is orthogonally invariant on both sides. Fix
any orthogonal $Q\in\real^{d\times d}$ whose first $k$ columns span the
unresolved subspace $U$ (such $Q$ exists because $U$ is $k$-dimensional).
Then $\tilde H_{e_i}:=H_{e_i}Q$ is again i.i.d.~$\mathcal{N}(0,1)$, and
\[
\norm{H_{e_i}P_U}_F^2 \;=\; \norm{\tilde H_{e_i} I_{k}'}_F^2 \;=\; \sum_{r=1}^{m}\sum_{s=1}^{k}\tilde H_{e_i,rs}^2,
\]
\[
\norm{H_{e_i}P_{U^\perp}}_F^2 \;=\; \sum_{r=1}^{m}\sum_{s=k+1}^{d}\tilde H_{e_i,rs}^2,
\]
where $I_k'$ selects the first $k$ columns. As sums of independent squared
standard normals over disjoint index sets, these are independent
$X_i\sim\chi^2_{mk}$ and $Y_i\sim\chi^2_{m(d-k)}$, and
$T_i:=X_i+Y_i=\norm{H_{e_i}}_F^2\sim\chi^2_{md}$.

The Beta--Gamma construction then gives $R_i:=X_i/T_i\sim\mathrm{Beta}(mk/2,m(d-k)/2)$,
independent of $T_i$: write $X_i=G_1/2$, $Y_i=G_2/2$ with $G_1\sim\Gamma(mk/2,1)$
and $G_2\sim\Gamma(m(d-k)/2,1)$ independent; then
$G_1/(G_1+G_2)\sim\mathrm{Beta}(mk/2,m(d-k)/2)\perp G_1+G_2$. The mean is
$\mathbb{E}[R_i]=\frac{mk/2}{md/2}=k/d$.

For the disagreement selector $D:=\arg\max_i T_i$, note that $(R_1,\ldots,R_n)\perp(T_1,\ldots,T_n)$
jointly: the candidate blocks $H_{e_i}$ are i.i.d., so the pairs
$(R_i,T_i)$ are independent across $i$, and within each pair
$R_i\perp T_i$ by the Beta--Gamma construction above. Therefore the full
$R$-family is independent of the full $T$-family, hence
$R_D\stackrel{d}{=}R_1\sim\mathrm{Beta}(mk/2,m(d-k)/2)$ and
$\mathbb{E}[R_D]=k/d$. \hfill$\square$

\subsection{Proof of Theorem~\ref{thm:high-d}}

Fix $n,m,k$ and let $d\to\infty$. By the above, $R_i\sim\mathrm{Beta}(mk/2,m(d-k)/2)$,
whose mass concentrates at $0$ as $d\to\infty$, so
$X_i/T_i\stackrel{p}{\to}0$. Hence $T_i$ is dominated by the resolved
energy $Y_i$ in the sense that
$T_i=Y_i(1+o_p(1))$. Therefore $D=\arg\max_i T_i$ and $\arg\max_i Y_i$
coincide with probability $\to 1$. $C=\arg\max_i X_i$ is, by rotational
invariance, independent of the $Y_i$. Hence
\[
\Pr(C=D)\to\Pr(C=\arg\max_i Y_i)=\sum_{i=1}^n\Pr(C=i)\Pr(\arg\max Y=i)=\tfrac{1}{n},
\]
so $\Pr(C\ne D)\to 1-1/n$. \hfill$\square$

\subsection{Proof of Proposition~\ref{prop:posterior}}

Under the local linear-Gaussian model, the unresolved coordinate
$z\in\real^{\dim(U_\tau)}$ has prior $\mathcal{N}(0,\Lambda_{\mathrm{cur}})$
and observation $y_e=H_e U_\tau z+\varepsilon_e$,
$\varepsilon_e\sim\mathcal{N}(0,\Sigma_e)$. The posterior is Gaussian with
precision
\[
\Lambda_e^{-1}
\;=\; \Lambda_{\mathrm{cur}}^{-1}\;+\;(H_e U_\tau)^\top \Sigma_e^{-1}(H_e U_\tau)
\;=\; \Lambda_{\mathrm{cur}}^{-1}+G_e,
\]
by the standard conjugate-Gaussian identity (precisions add). Taking the
inverse yields $\Lambda_e=(\Lambda_{\mathrm{cur}}^{-1}+G_e)^{-1}$, and
applying the Woodbury identity,
$\Lambda_e=\Lambda_{\mathrm{cur}}-\Lambda_{\mathrm{cur}}(\Lambda_{\mathrm{cur}}+G_e^{-1})^{-1}\Lambda_{\mathrm{cur}}$
when $G_e$ is invertible, exposing the posterior-covariance \emph{reduction}
that appears in \eqref{eq:aopt-score}. \hfill$\square$

\subsection{Proof of Proposition~\ref{prop:fisher}}

For isotropic $\Sigma_e=\sigma_e^2 I$,
$G_e=\sigma_e^{-2}U_\tau^\top H_e^\top H_e U_\tau$. Taking trace and using
$\tr(A^\top A)=\norm{A}_F^2$, $\tr(G_e)=\sigma_e^{-2}\norm{H_e U_\tau}_F^2$.
\hfill$\square$

\subsection{Proof of Corollary~\ref{cor:k1}}

When $\dim(U_\tau)=1$, $\Lambda_{\mathrm{cur}}$ is a positive scalar
$\lambda$ and $G_e$ is a non-negative scalar $g_e$. Posterior variance is
$\lambda/(1+\lambda g_e)$, monotone decreasing in $g_e$. Under isotropic
noise, $g_e\propto\norm{H_e U_\tau}_F^2=\scorecart(e)$ by
\cref{prop:fisher}. Hence minimizing posterior variance $\Leftrightarrow$
maximizing $\scorecart$. \hfill$\square$

\subsection{Proof of Proposition~\ref{prop:eig}}

Under the local linear-Gaussian model, prior and posterior on the
unresolved coordinates are Gaussian with covariances $\Lambda_{\mathrm{cur}}$
and $\Lambda_e=(\Lambda_{\mathrm{cur}}^{-1}+G_e)^{-1}$ (\cref{prop:posterior}).
Expected information gain is the expected KL divergence from prior to
posterior \citep{lindley1956,chaloner1995review}; for $d$-dimensional
Gaussians $\mathcal{N}(\mu_0,\Sigma_0)\to\mathcal{N}(\mu_1,\Sigma_1)$,
$\mathrm{KL}=\tfrac{1}{2}[\tr(\Sigma_0^{-1}\Sigma_1)-d+(\mu_1-\mu_0)^\top\Sigma_0^{-1}(\mu_1-\mu_0)-\log\det(\Sigma_0^{-1}\Sigma_1)]$.
Expectation over the marginal of $y_e$ kills the data-dependent terms in
the Gaussian linear case \citep[Eq.~7]{chaloner1995review} and yields
$\EIG(e)=\tfrac{1}{2}\log\det(\Lambda_{\mathrm{cur}}\Lambda_e^{-1})$.
Substituting the precision identity,
\[
\EIG(e)=\tfrac{1}{2}\log\det(I+\Lambda_{\mathrm{cur}}G_e)
=\tfrac{1}{2}\log\det(I+\Lambda_{\mathrm{cur}}^{1/2}G_e\Lambda_{\mathrm{cur}}^{1/2}),
\]
the latter by Sylvester's determinant identity
$\det(I+AB)=\det(I+BA)$. Expanding in the small-information regime via
$\log\det(I+A)=\tr(A)-\tfrac{1}{2}\tr(A^2)+O(\norm{A}_F^3)$ (valid for
$\norm{A}<1$),
\[
\EIG(e)=\tfrac{1}{2}\tr(\Lambda_{\mathrm{cur}}G_e)+O\!\big(\norm{\Lambda_{\mathrm{cur}}G_e}_F^2\big).
\]
Under isotropic prior and isotropic noise, $\Lambda_{\mathrm{cur}}\propto I$
and $G_e\propto H_e U_\tau^\top H_e U_\tau$, so the linear term collapses
to $\propto\scorecart(e)$. \hfill$\square$

\subsection{Proof of Proposition~\ref{prop:boxhill}}

The Box--Hill criterion between two candidate predictive distributions
$\mathcal{N}(m_i, V_i)$ is the symmetrized KL divergence
$\tfrac{1}{2}(\mathrm{KL}(p_1\|p_2)+\mathrm{KL}(p_2\|p_1))$, which for
Gaussians yields \eqref{eq:bh} \citep{boxhill1967}. When the two
predictive covariances are equal isotropic, the trace terms collapse to a
constant and the discriminative term becomes
$\sigma_e^{-2}\norm{\Delta}_2^2$. In the linearized model around the
current posterior mean, $\Delta$ is precisely the row-block of $H_e$
corresponding to the model-pair disagreement; maximizing
$\sigma_e^{-2}\norm{\Delta}_2^2$ across candidates is the
disagreement-magnitude heuristic, which by \cref{thm:useful-frac} captures
a $k/d$ fraction of the unresolved information. The linear-Gaussian
Box--Hill therefore reduces to disagreement in this isotropic limit. The
non-collapsed Box--Hill baseline used in \cref{sec:cascade} keeps
model-specific predictive covariances, which is why its empirical curve need
not coincide with disagreement. \hfill$\square$

\section{Worked Example of the Disagreement Operator}
\label{app:he-example}

Consider $L=3$ library members, $n_e=2$ observation times, and
$|C|=2$ controversial coordinates. Let the local Jacobians be
\[
J_{1,e},J_{2,e},J_{3,e}\in\real^{2\times 2}.
\]
The three pairwise disagreement blocks are
\[
D_{12,e}=J_{1,e}-J_{2,e},\quad
D_{13,e}=J_{1,e}-J_{3,e},\quad
D_{23,e}=J_{2,e}-J_{3,e},
\]
each in $\real^{2\times 2}$. Stacking them gives
\[
H_e=\begin{bmatrix}
D_{12,e}\\
D_{13,e}\\
D_{23,e}
\end{bmatrix}\in\real^{6\times 2}.
\]
For a controversial-coordinate perturbation $z\in\real^2$, the vector
$H_e z\in\real^6$ contains the first-order pairwise predictive differences
across the three model pairs and the two observation times. The same
construction is used in the implementation: every experiment contributes one
block $H_e$, and the accumulated disagreement matrix
$H_{\mathrm{cur}}$ is the vertical concatenation of the executed blocks.

\section{Closed-Form BOED Baseline Implementations}
\label{app:baselines}

We give the closed-form baselines we use in the cascade benchmark. All
operate inside the same local linear-Gaussian framework as the analysis in
\cref{sec:boed}.

\paragraph{Expected information gain (EIG).}
Under \cref{prop:posterior,prop:eig}, the closed-form EIG score is
\begin{equation}
\mathrm{EIG}(e) = \tfrac{1}{2}\log\det\!\big(I+\Lambda_{\mathrm{cur}}^{1/2}G_e\Lambda_{\mathrm{cur}}^{1/2}\big),
\end{equation}
which costs one $k\times k$ log-determinant per candidate.

\paragraph{Box--Hill.}
Under \cref{prop:boxhill} we use
\begin{equation}
\mathrm{BH}(e) = \sum_{i<j}\tfrac{1}{2}\Delta_{ij}^\top(V_i^{-1}+V_j^{-1})\Delta_{ij}
\end{equation}
over all library pairs $(i,j)$, with $\Delta_{ij}=m_i(e)-m_j(e)$ and
$V_i$ the predictive covariance under model $i$ evaluated at $e$. In the
isotropic shared-noise limit, this reduces to
$\sigma_e^{-2}\sum_{i<j}\norm{\Delta_{ij}}^2$, i.e.~the disagreement-magnitude
baseline. The cascade Box--Hill curve in the main paper uses the full
model-specific $V_i$, so it is not expected to collapse to disagreement.

\paragraph{Local T-optimal.}
Local T-opt \citep{atkinsonfedorov1975,pukelsheim2006optimal} selects
experiments that maximize the sum-of-squared predictive differences
between the two best-fit library members around the current parameter
estimate. Under the local linear-Gaussian model it coincides with the
$(i,j)$-restricted Box--Hill criterion at a single dominant pair.

\paragraph{Why \method\ is not strictly dominated by these baselines.}
EIG and Box--Hill both require predictive distributions of \emph{every}
library member at \emph{every} candidate experiment. \method\ requires
only the disagreement design $H_e$ and the current unresolved basis
$U_\tau$, so it continues to be well-defined under libraries that are
specified as simulators without tractable predictive distributions. This
is a common deployment profile for AI-scientist stacks whose library is a
set of heterogeneous simulators or black-box models.

\section{Cascade Benchmark Details}
\label{app:cascade}

The cascade system is a sequence of first-order ODEs with a
scalable-dimension mechanism basis: $d$ kinetic coefficients plus a small
number of output-coupling mechanisms, following the general template for
model-discrimination experiments in biochemical networks
\citep{vanhof2014optimal}. Library members are obtained by
omitting one or two kinetic coefficients from the shared basis; the
controversial component $a^{\star}_C$ is the vector of omitted
coefficients. Experiments correspond to forcing profiles with different
time-to-peak, amplitude, and dwell combinations. Local Jacobian blocks are
computed from ODE sensitivities in closed form.

\paragraph{Protocol.}
For each $d\in\{2,4,8,16\}$: $6$ truths (different realizations of
$a^{\star}_C$) $\times$ $24$ noise seeds $=144$ trials. Rounds $=3$.
Candidates $=8$ forcing profiles.

\paragraph{Full results.}
\Cref{tab:cascade-full} reproduces the main robustness table with
disagreement included and with standard deviations added. Runtime per
selection step in \cref{tab:runtime-cascade}, Appendix~\ref{app:runtimes}.

\begin{table*}[t]
\caption{Cascade robustness full table. Regret is averaged across
$144$ trials; values in parentheses are $\pm 1$ standard deviation across
trials.}
\label{tab:cascade-full}
\centering
\small
\begin{tabular}{r c c c c c c c}
\toprule
$d$ & Init.~unres.~dim & Raw HB & \uopt\ HB & Disagree HB & Raw regret & \uopt\ regret & Disagree regret \\
\midrule
2  & 1  & 0.44 & 0.44 & 0.44 & 0.05 & 0.05 & 0.05 \\
4  & 3  & 0.00 & 0.09 & 0.00 & 0.31 & 0.42 & 0.31 \\
8  & 6  & 0.02 & 0.65 & 0.02 & 19.94 & 0.01 & 19.94 \\
16 & 12 & 0.07 & 0.72 & 0.07 &  2.83 & 0.01 &  2.83 \\
\bottomrule
\end{tabular}
\end{table*}

\section{PK Divergence Details}
\label{app:pk-detail}

\begin{table}[h]
\caption{Per-truth rounds to identification. Raw \method\ and \uopt\ are
identical on all seven PK truths; both beat disagreement only on
\emph{absorption\_variant} (bold). Random E[rounds] is a Monte Carlo
estimate over $10{,}000$ permutations.}
\centering
\small
\begin{tabular}{l c c c c c}
\toprule
Truth & Oracle & Raw & \uopt\ & Disagree & Random \\
\midrule
\textbf{absorption\_variant}   & B & \textbf{1} & \textbf{1} & \textbf{2} & 4.00 \\
absorption\_variant\_slow      & B & 1 & 1 & 1 & 1.14 \\
distribution\_variant\_easy    & C & 0 & 0 & 0 & 0.00 \\
distribution\_variant\_hard    & C & 1 & 1 & 1 & 1.60 \\
distribution\_variant\_subtle  & C & 1 & 1 & 1 & 2.40 \\
mixed\_absorption              & A & 2 & 2 & 2 & 5.33 \\
mixed\_balanced                & A & 2 & 2 & 2 & 5.33 \\
\bottomrule
\end{tabular}
\end{table}

\paragraph{Method sequences.}
Raw \method: $[E_1, E_2, E_4, E_5, E_3, E_7, E_6]$;
\uopt: $[E_1, E_2, E_3, E_4, E_5, E_6, E_7]$;
disagreement: $[E_2, E_1, E_6, E_7, E_5, E_3, E_4]$.
All structured methods dominate random.

\begin{figure}[h]
\centering
\safeincludegraphics[width=0.7\linewidth]{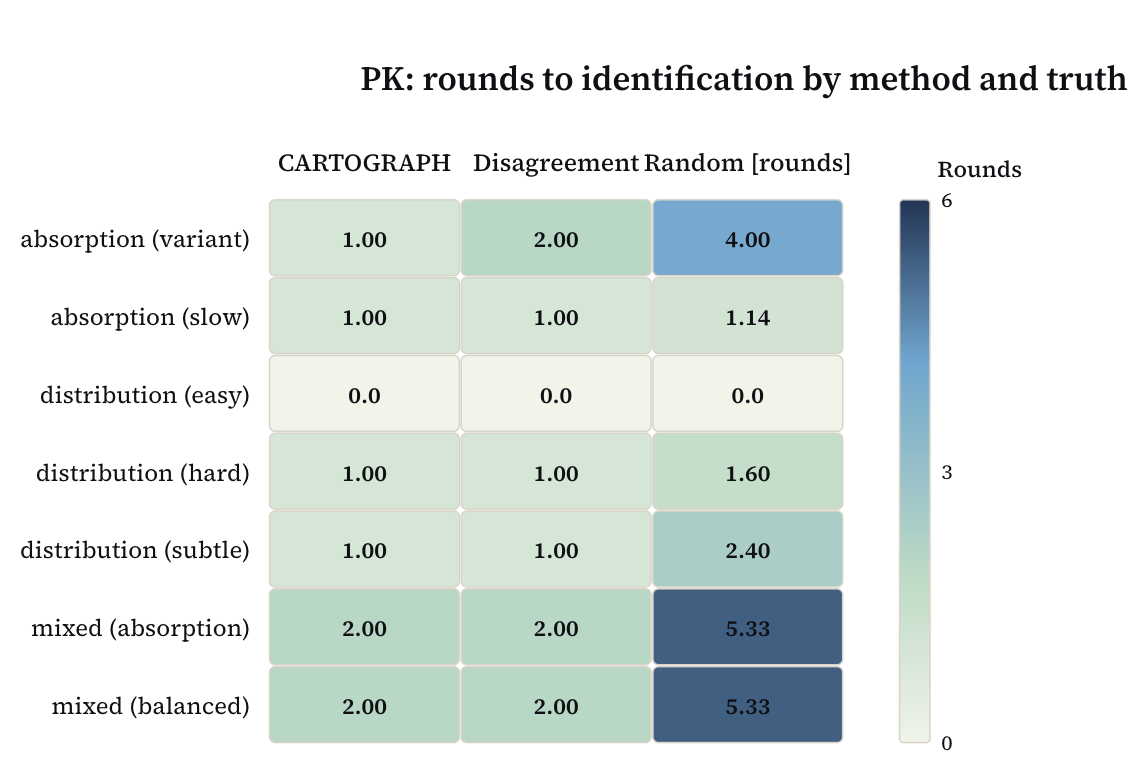}
\caption{PK rounds to identification, per truth. Raw \method\ and \uopt\
identify in the same round on all seven truths; the structured methods
collectively differ from disagreement on one case
(\emph{absorption\_variant}).}
\label{fig:pk-rounds}
\end{figure}

\section{Scaling Experiment Full Tables}
\label{app:scaling-tables}

\begin{table*}[h]
\caption{Scaling experiment, $k\!=\!1$, $500$ random-candidate instances
per $d$. ``Useful Frac'' is the empirical fraction of disagreement-selected
energy that falls on unresolved directions; ``Theory $k/d$'' is the exact
prediction of \cref{thm:useful-frac}.}
\centering
\small
\begin{tabular}{r c c c c c c}
\toprule
$d$ & Disagree rate & CART win (proj) & CART win (rank) & Overall adv (rank) & Useful Frac & Theory $k/d$ \\
\midrule
 2 & 54.8\% & 100.0\% & 76.3\% & 41.8\% & 0.501 & 0.500 \\
 3 & 63.6\% & 100.0\% & 71.7\% & 45.6\% & 0.323 & 0.333 \\
 4 & 67.2\% & 100.0\% & 68.2\% & 45.8\% & 0.242 & 0.250 \\
 6 & 70.0\% & 100.0\% & 61.4\% & 43.0\% & 0.175 & 0.167 \\
 8 & 73.4\% & 100.0\% & 61.6\% & 45.2\% & 0.196 & 0.125 \\
10 & 75.0\% & 100.0\% & 70.9\% & 53.2\% & 0.175 & 0.100 \\
12 & 76.2\% & 100.0\% & 67.7\% & 51.6\% & 0.170 & 0.083 \\
15 & 73.0\% & 100.0\% & 72.6\% & 53.0\% & 0.165 & 0.067 \\
\bottomrule
\end{tabular}
\end{table*}

\begin{table*}[h]
\caption{Scaling experiment, $k\!=\!2$, $500$ random-candidate instances
per $d$.}
\centering
\small
\begin{tabular}{r c c c c c c}
\toprule
$d$ & Disagree rate & CART win (proj) & CART win (rank) & Overall adv (rank) & Useful Frac & Theory $k/d$ \\
\midrule
 3 & 41.2\% & 100.0\% & 68.4\% & 28.2\% & 0.672 & 0.667 \\
 4 & 55.2\% & 100.0\% & 65.2\% & 36.0\% & 0.501 & 0.500 \\
 6 & 59.8\% & 100.0\% & 64.9\% & 38.8\% & 0.340 & 0.333 \\
 8 & 69.0\% & 100.0\% & 63.2\% & 43.6\% & 0.271 & 0.250 \\
10 & 69.6\% & 100.0\% & 69.5\% & 48.4\% & 0.238 & 0.200 \\
12 & 71.4\% & 100.0\% & 65.5\% & 46.8\% & 0.220 & 0.167 \\
15 & 72.6\% & 100.0\% & 70.5\% & 51.2\% & 0.200 & 0.133 \\
\bottomrule
\end{tabular}
\end{table*}

\begin{figure}[h]
\centering
\safeincludegraphics[width=0.7\linewidth]{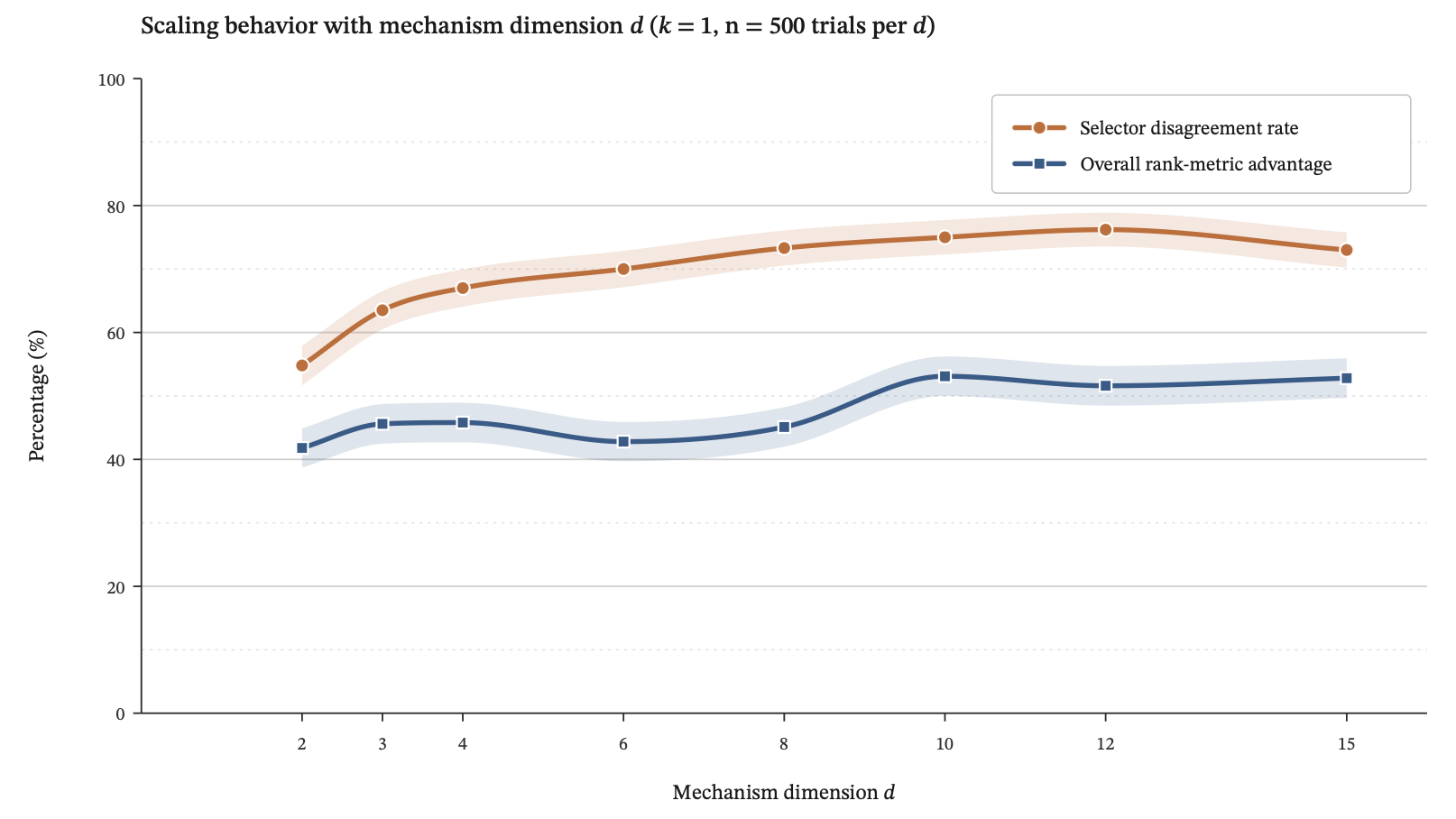}
\caption{Scaling curves. At low $d$ disagreement and projection nearly
agree; at high $d$ the projection selector's pointwise-optimal behavior
under the unresolved-energy metric emerges.}
\label{fig:scaling-curves}
\end{figure}

\section{EPA Real-Data Full Breakdown}
\label{app:epa}

We evaluated oral series from the EPA CvTdb database
\citep{sayre2020cvtdb}. The cohort filter was: oral route, $\ge 10$
concentration measurements, maximum time $\le 24\,\mathrm{h}$, one series
per chemical. The library is $\{A:\text{1-compartment oral}, B:
\text{delayed-absorption oral}, C:\text{2-compartment oral}\}$.

\paragraph{Cohort-level summary.}

\begin{table}[h]
\caption{EPA cohort-level summary for raw \method\ (BIC gap margin $=2.00$).}
\centering
\small
\begin{tabular}{l c c c c c}
\toprule
Split & Raw \method\ margin & Disagree margin & T-opt margin & Random E[margin] & Raw \method\ hit \\
\midrule
All series ($n=8$)   &  7.70 & 7.69 & 8.58 & 4.29 & 50.0\% \\
Active subset ($n=5$) & 13.55 & 13.53 & 13.55 & 7.99 & 80.0\% \\
\bottomrule
\end{tabular}
\end{table}

\paragraph{Pairwise results.}
All series: raw \method\ vs Disagree $1\mathrm{W}/7\mathrm{T}/0\mathrm{L}$;
raw \method\ vs T-opt $0\mathrm{W}/7\mathrm{T}/1\mathrm{L}$; T-opt vs Disagree
$2\mathrm{W}/6\mathrm{T}/0\mathrm{L}$.

\paragraph{Per-series.}
Dichloromethane, 1,2-Dichloroethane, Trichloroethylene, Benzo[a]pyrene,
Chloroform, Methyl tert-butyl ether, Valproic acid, Glycine. On three of
these ($n_{\mathrm{succ}}=3$) the unresolved-dim is $0$ already at the
warm start: every acquisition rule must tie. On the other five,
raw \method\ matches T-opt on four and trails on one.

\paragraph{Fit failures.}
The cohort contained $96$ oral series; $88$ failed fit in at least one
library member and were skipped. These fit failures are themselves
informative: they indicate that the three-model library is inadequate for
a large fraction of EPA chemicals, which is exactly the situation the
\emph{refuse} layer is designed to surface.

\section{A-Lab Retrospective Audit Details}
\label{app:alab}

We downloaded the corrected public A-Lab supplementary data
\citep{szymanski2023alab,szymanski2026correction}, including the synthesis
results CSV and `Refinement-Table.xlsx'. The audit population is the $40$
originally positive claims labelled `Success' or `Partial' in the synthesis
CSV. External labels come from the corrected manual conclusions: only
structure or composition marked inconclusive is counted as `inconclusive'.
We exclude ordering ambiguity alone because it is a crystallographic
refinement issue orthogonal to whether the target phase was synthesized.

For each claim we compute a materials-domain residual
\[
\rho_{\mathrm{A\text{-}Lab}}
=\sqrt{(R_{\mathrm{wp}}/20)^2
       + ((100-w_{\mathrm{target}})/100)^2
       + (w_{\mathrm{alt}}/100)^2},
\]
where $R_{\mathrm{wp}}$ is the manual-refinement weighted-profile residual,
$w_{\mathrm{target}}$ is the target phase fraction, and $w_{\mathrm{alt}}$
is the largest non-target phase fraction. We calibrate $\delta$ as the
$95$th percentile of $\rho_{\mathrm{A\text{-}Lab}}$ on confirmed `Success'
rows, yielding $\delta=0.776$, then freeze it before evaluating the full
positive-claim set. A deterministic bootstrap over the calibration rows
($2000$ resamples, seed $0$) gives a broad but transparent calibration
interval, $\delta\in[0.496,1.088]$, reflecting the small public cohort.

\begin{table}[h]
\caption{A-Lab retrospective audit. The guard flags all four
post-correction inconclusive positive claims, at the cost of flagging four
confirmed but complex multiphase positive claims for human review.}
\centering
\small
\begin{tabular}{l c c c}
\toprule
External label & Claims & Passed & Flagged \\
\midrule
Confirmed & 36 & 32 & 4 \\
Inconclusive & 4 & 0 & 4 \\
\bottomrule
\end{tabular}
\end{table}

The four inconclusive claims flagged are CaGd$_2$Zr(GaO$_3$)$_4$,
KBaGdWO$_6$, Mn$_7$(P$_2$O$_7$)$_4$, and Mg$_3$MnNi$_3$O$_8$. An Rwp-only
baseline calibrated by the same $95$th-percentile protocol flags $0/4$
inconclusive claims and $2/36$ confirmed claims. A target-deficit-only
baseline flags $4/4$ inconclusive and $4/36$ confirmed claims, matching the
combined guard's confirmed flag count but without the refinement-residual
audit trail. We therefore use this result narrowly: it shows that a published
autonomous-discovery output can be converted into an auditable pass/flag
log under a fixed calibration rule, not that \method\ independently
settles the crystallographic status of the A-Lab targets.

\section{Per-Feature Refusal Diagnostics}
\label{app:refuse-features}

\begin{table}[h]
\caption{Per-feature residual breakdown. Higher $=$ the feature detects
more misfit. $C_{\max}$ is the dominant detector for all three failure
mechanisms; the control's $C_{\max}$ residual is much lower.}
\centering
\small
\begin{tabular}{l c c c c}
\toprule
Feature        & Time-var & Saturable & Enterohep & Control \\
\midrule
$T_{\max}$        & 0.00 & 0.00 & 0.00 & 0.00 \\
$C_{\max}$        & 1.83 & 1.36 & 1.74 & 0.73 \\
$\mathrm{AUC}_{\mathrm{frac}}$ & 0.06 & 0.08 & 0.11 & 0.03 \\
Terminal slope & 0.22 & 0.38 & 0.29 & 0.01 \\
Loglin RMSE    & 0.25 & 0.21 & 0.22 & 0.07 \\
\bottomrule
\end{tabular}
\end{table}

\section{Failure-Mode Catalog for \method}
\label{app:failure-modes}

We catalog conditions under which \method\ is expected to be wrong.

\textbf{1.~Non-linear local regime.}
\cref{prop:fisher,prop:eig,prop:boxhill} rely on the local linear-Gaussian
approximation. When the system's curvature around the current posterior
mean is large, the first-order equivalences degrade. Mitigation: score with
the noise-weighted exact form \eqref{eq:aopt-score}, which remains valid
locally even when the first-order approximations drift, and re-linearize
after each experiment.

\textbf{2.~Heavy-tailed or adversarial noise.}
The Gaussian assumption underwrites the closed-form EIG. Under heavy tails
or structured model-dependent noise, $\Sigma_e$ mis-specifies the true
noise and $G_e$ is off. Mitigation: use the noise-weighted form with an
empirical Bayes estimate of $\Sigma_e$.

\textbf{3.~Prior that excludes the truth (refuse failure by miscalibration).}
If $\delta$ is miscalibrated high, \method\ identifies out-of-library
mechanisms. This is detectable post-hoc from the
$(\rho_t,\mathrm{outcome})$ log: sustained increases in $\rho$ under
accumulating evidence are an audit signal.

\textbf{4.~Degenerate unresolved subspace at warm start.}
If the warm-start design is rich enough that $\dim(U_\tau)=0$ already,
every rule ties and \method\ reduces to library identification. We flag
this at the start of the loop and return a ``no-op'' status.

\textbf{5.~Near-isotropy of the posterior on $U_\tau$.}
When $\Lambda_{\mathrm{cur}}\propto I$, $\uopt$ and raw \method\ nearly
coincide (first-order equivalence). The cascade $d\!=\!2$ tie regime is
exactly this case. No selection advantage is possible and reporting
near-ties is the honest output.

\section{Runtime Table}
\label{app:runtimes}

\begin{table}[h]
\caption{Per-step runtime in milliseconds, cascade benchmark, single CPU
core.}
\label{tab:runtime-cascade}
\centering
\small
\begin{tabular}{r c c c}
\toprule
$d$ & Raw score (ms) & \uopt\ score (ms) & Disagree score (ms) \\
\midrule
 2 & 0.006 & 0.034 & 0.0003 \\
 4 & 0.006 & 0.039 & 0.0004 \\
 8 & 0.015 & 0.268 & 0.0004 \\
16 & 0.032 & 1.299 & 0.0005 \\
\bottomrule
\end{tabular}
\end{table}

\section{Worked LLM-in-the-Loop Example}
\label{app:llm-agent}

We show how \method\ plugs into an LLM-based AI scientist. The working
example is the pharmacokinetic library from \cref{sec:pk}; the LLM role is
\emph{candidate proposal}, not \emph{decision}.

\paragraph{Setting.}
The library is $\{A, B, C\}$: one-compartment oral, delayed-absorption
oral, and two-compartment oral models, following standard pharmacokinetic
parameterizations \citep{gabrielsson2016pkbook}. The LLM is handed the library cards
and a natural-language description of the current state of the loop.

\paragraph{LLM prompt (abbreviated).}
``You are an experimental planner for a pharmacokinetic library containing
one-compartment oral, delayed-absorption oral, and two-compartment oral
models. Candidate experiments differ in route (oral or IV), dose, sampling
schedule (early-dense, mid-dense, late-dense), and observation horizon.
Propose up to five candidate experiments as JSON objects with fields
\texttt{route}, \texttt{dose\_mg\_per\_kg}, \texttt{sampling\_profile},
\texttt{horizon\_h}, \texttt{rationale}. Only propose feasible experiments.''

\paragraph{LLM output (one run, abbreviated).}
\begin{quote}\small\ttfamily
[\{"route":"oral","dose":5,"sampling":"early\_dense","horizon":6,
"rationale":"constrain $k_a$"\},
\{"route":"oral","dose":5,"sampling":"late\_dense","horizon":24,
"rationale":"constrain terminal slope"\},
\{"route":"IV","dose":2,"sampling":"early\_dense","horizon":4,
"rationale":"isolate distribution vs absorption"\},
\{"route":"oral","dose":10,"sampling":"mid\_dense","horizon":12,
"rationale":"resolve A vs C on $C_{\max}$"\},
\{"route":"IV","dose":2,"sampling":"mid\_dense","horizon":12,
"rationale":"resolve B vs C on late phase"\}]
\end{quote}

\paragraph{\method's role.}
Each proposal is lowered into a candidate disagreement block $H_e$ by
evaluating the library's feature-bridge predictions under each candidate
experiment. \method\ scores them via \eqref{eq:raw-score} or
\eqref{eq:aopt-score}, picks $e^\star$, and returns two structured
objects: (i)~a scalar $\scorecart(e^\star)$ for the planner, and (ii)~the
current unresolved dimension $\dim(U_\tau)$. The planner uses these two
numbers to decide whether to ask the LLM for more candidates (if
$\scorecart$ is small but $\dim(U_\tau)>0$) or to execute $e^\star$ and
continue the loop.

\paragraph{What this demonstrates.}
The LLM plays a role it is good at (generating plausible candidates with
natural-language rationale) and \emph{not} a role it is bad at
(discriminating candidates under a joint posterior). \method\ supplies the
latter. Combined with the refuse layer, the resulting agent has three
behaviors no LLM-only planner produces natively: it refuses identification
when the library is structurally wrong, it revokes tentative
identifications under contradictory evidence, and it certifies when the
unresolved question has been closed.

\end{document}